\documentclass[11pt]{article}

\usepackage[preprint]{acl}

\usepackage{times}
\usepackage{latexsym}

\usepackage{hyperref}
\usepackage{url}
\usepackage{xcolor}
\usepackage{booktabs}
\usepackage{multirow}
\usepackage{amssymb}
\usepackage{tabularx}
\usepackage{tcolorbox}
\usepackage{subcaption}
\usepackage{colortbl}
\usepackage{makecell}
\usepackage{amsmath}
\usepackage{cleveref}
\usepackage{enumitem}
\usepackage[T1]{fontenc}
\usepackage[utf8]{inputenc}

\usepackage{microtype}

\usepackage{inconsolata}

\usepackage{graphicx}

\title{SALT: Step-level Advantage Assignment for Long-horizon Agents \\ via Trajectory Graph}
\author{Jiazheng Li\textsuperscript{1}\thanks{Work done during internship at Amazon.}, Yawei Wang\textsuperscript{2}, David Yan\textsuperscript{2}, Yijun Tian\textsuperscript{2} \\  {\bf Zhichao Xu\textsuperscript{2}, Huan Song\textsuperscript{2}, Panpan Xu\textsuperscript{2}\thanks{Corresponding author: \texttt{xupanpan@amazon.com}}, Lin Lee Cheong\textsuperscript{2}}\\ 
$^1$University of Connecticut \quad
$^2$Amazon \\
\texttt{jiazheng.li@uconn.edu} \qquad
\texttt{\{yawenwan, qiaojiny, yijunt}\\
\texttt{xzhichao, huanso, xupanpan, lcheong\}@amazon.com}
}

\begin{document}

\maketitle

\begin{abstract}
Large Language Models (LLMs) have demonstrated remarkable capabilities, enabling language agents to excel at single-turn tasks. However, their application to complex, multi-step, and long-horizon tasks remains challenging. While reinforcement learning (RL) offers a promising avenue for addressing these challenges, mainstream approaches typically rely solely on sparse, outcome-based rewards, a limitation that becomes especially problematic for group-based RL algorithms lacking critic models, such as Group Relative Policy Optimization (GRPO). In such methods, uniformly rewarding or penalizing all actions within a trajectory can lead to training instability and suboptimal policies, because beneficial and detrimental actions are often entangled across multi-step interactions.
To address this challenge, we propose \textbf{SALT}, a novel and lightweight framework that provides a finer-grained advantage assignment, derived solely from outcome rewards. We achieve this by constructing a graph from trajectories of the same prompt, which allows us to quantify the quality of each step and assign advantages accordingly.
Crucially, SALT is designed as a plug-and-play module that seamlessly integrates with existing group-based RL algorithms, requiring no modifications to the rollout procedure and introducing negligible computational overhead.
Extensive experiments on the WebShop, ALFWorld, and AppWorld benchmarks with various model sizes demonstrate that SALT consistently improves performance.
We also conduct a thorough analysis to validate the design choices behind SALT and offer actionable insights.
\end{abstract}

\section{Introduction}
\begin{figure}[t]
  \centering
  \includegraphics[width=0.9\columnwidth]{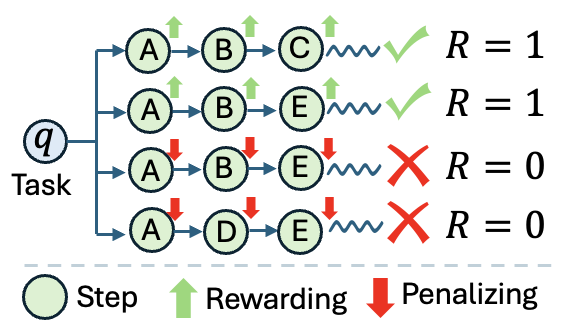}
  \caption{In group-based agentic RL, steps like A, B, or E that appear across multiple trajectories sometimes receive inconsistent advantages, being rewarded in some and penalized in others. This inconsistency stems from trajectory-level reward assignment and can lead to gradient conflicts during the policy update process.}
  \label{fig:conflicts}
  \vspace{-0.3cm}
\end{figure}
Recent advances in large language models (LLMs) have demonstrated their remarkable potential to function as intelligent agents, enabling a diverse array of applications, including web agents~\citep{wu2025webdancer,li2025websailor}, search agents~\citep{jin2025search,sun2025zerosearch}, coding agents~\citep{deepswe2025,yang2024swe}, and embodied agents~\citep{xi2024agentgym,intelligence2025pi_}. While LLM-based agents excel at simple, single-step tasks, like weather forecast inquiry with API calling, they often struggle in complex, long-horizon scenarios that require sustained, multi-step interaction with external environments~\citep{zhou2025sweet, loop}. For example, in the AppWorld benchmark~\citep{appworld}, completing the task \textit{Like all the Venmo transactions from today involving any of my roommates on my Venmo social feed} requires GPT-4o~\citep{gpt4o} using the ReAct framework~\citep{yao2023react} to execute a sequence of 18 precisely coordinated steps, from \textit{logging in} to \textit{identifying transactions} and finally \textit{liking them}, which makes them particularly challenging for LLM-based agents.

To tackle these challenges, recent work has increasingly turned to reinforcement learning~\citep[RL,][]{wang2025ui, xi2025agentgym}, particularly in goal-oriented settings where collecting high-quality expert demonstrations for supervised fine-tuning (SFT) is not only labor-intensive but also limits generalization. In contrast, RL directly optimizes for verifiable objectives, such as task success rate, and naturally fosters the “explore-and-exploit” behavior essential for navigating dynamic, interactive environments~\citep{chen2025r1-code,singh2025agentic}. 
Among RL approaches, group-based algorithms, notably Group Relative Policy Optimization~\citep[GRPO,][]{grpo}, have gained widespread adoption for training LLM agents across diverse domains, including search~\citep{jin2025search}, tool use~\cite{singh2025agentic}, and more. Their appeal lies in their simplicity and scalability compared to non-group-based methods like PPO~\citep{ppo}, which requires maintaining a separate critic model to estimate value functions.

\begin{figure*}[t]
  \includegraphics[width=\textwidth]{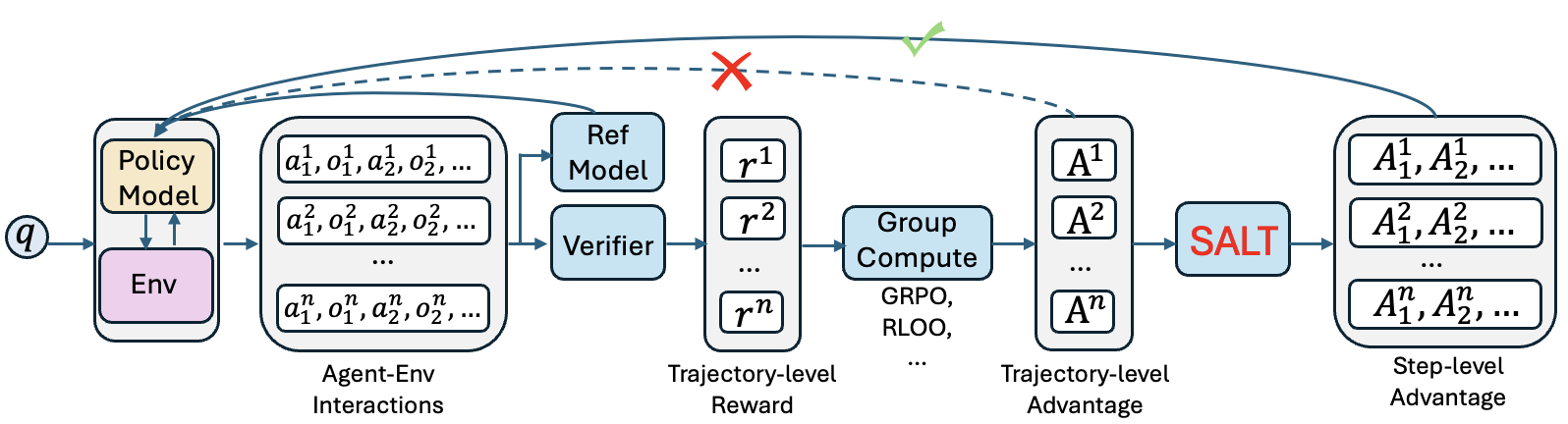}
  \caption{\textbf{Training pipeline with SALT}. After parallel rollouts and reward assignment, group-based RL computes trajectory-level advantages. SALT is then inserted and refines these into step-level advantages by leveraging cross-trajectory structure, enabling fine-grained policy updates, without altering the rollout or reward pipeline.}
  \label{fig:overview}
  \vspace{-0.3cm}
\end{figure*}

Group-based RL algorithms typically compute advantages by comparing the relative final rewards of multiple trajectories that attempt the same task, then uniformly broadcasting this scalar advantage to every step within each trajectory. Existing works~\citep{gigpo,zhang2025rlvmr,zhou2025sweet} have shown that this coarse-grained credit assignment is fundamentally suboptimal, especially in long-horizon settings where beneficial and detrimental actions are often entangled. Rewarding or penalizing all actions based solely on final outcomes can lead to unstable policy updates and degraded performance.

To mitigate the limitation of coarse-grained advantage assignment, we introduce SALT, a lightweight yet highly effective mechanism that delivers fine-grained, step-level advantages. SALT is motivated by a simple observation: \emph{trajectories that solve the same task often share some steps yet diverge at others.}
We believe that steps shared across all trajectories are typically neutral or non-differentiating (e.g., Step A in Figure~\ref{fig:conflicts}), while those unique to high-reward trajectories are likely beneficial (e.g., Step C), and those exclusive to failures are likely detrimental (e.g., Step D).

Unlike prior methods, SALT explicitly identifies shared versus distinct steps through trajectory graph construction, requiring no additional supervision or reward models. It operates solely on sparse, episodic returns. Concretely, given a task, SALT first generates a batch of parallel rollouts, as in standard group-based RL. It then unifies these trajectories into a single trajectory graph using two elementary operations: \textit{merge} and \textit{diverge}. This structure enables quantitative step-level advantage refinement, which directly guides policy updates. Designed as a plug-and-play module, SALT integrates seamlessly into existing pipelines, inheriting their algorithmic strengths while adding negligible computational overhead.

% Thus, unlike prior works, by explicitly identifying shared versus distinct steps across trajectories, SALT refines per-step advantage assignment — without requiring any additional supervision, reward models, or environment instrumentation, solely based on the final episodic returns.

% Concretely, given a task, we first roll out a batch of trajectories in parallel, following standard group-based RL practice. We then construct these separate trajectories into a single trajectory graph via two elementary operations: \textit{diverge} and \textit{merge}. This graph structure enables us to quantitatively assess the quality of each step by updating its advantage score which is furthur used to guide the policy update. Moreover, SALT is designed as a plug-and-play module, seamlessly integrable into existing group-based RL pipelines. It inherits the latest algorithmic advances while introducing only negligible computational overhead.
 
% Thus, leveraging solely on the final episodic returns and without introducing any additional signals, SALT translates these patterns into precise, step-level advantages, enabling more accurate advantage assignment and faster learning. 

We evaluate SALT on three challenging long-horizon benchmarks: ALFWorld (embodied reasoning), WebShop (interactive e-commerce), and AppWorld (digital personal assistance). By plugging SALT into GRPO and RLOO, we obtain consistent performance gains across model scales (1.5B, 7B and 32B parameters).

In summary, our contributions are as follows:

\begin{enumerate}
    \item We propose SALT, a novel framework for step-level advantage assignment in long-horizon agentic RL. By constructing a trajectory graph to distinguish shared and distinct steps, SALT produces fine-grained advantages without any additional supervision or reward models.
    \item SALT is a lightweight, plug-and-play module that integrates effortlessly into existing group-based RL pipelines. It consistently enhances performance while incurring negligible computational cost.
    \item Through extensive experiments on ALFWorld, WebShop, and AppWorld, we demonstrate SALT’s consistent superiority across various tasks and model sizes. Detailed analysis and case studies further validate the effectiveness and interpretability of our approach.
\end{enumerate}

\section{Preliminaries}
In this section, we formalize the problem setting and review the fundamental reinforcement learning algorithms relevant to our approach.
\subsection{LLM Agents}
\label{sec:pre_agents}
We formalize the interaction between an agent and its environment in long-horizon tasks as a Markov Decision Process (MDP). While the classical MDP assumes full observability of the environment state, in the context of LLM agents, the agent typically receives information through natural language observations rather than direct access to the underlying state. To accommodate this, we adopt a formulation that explicitly distinguishes between the environment’s true state and the agent’s observation. Formally, we define the process as a tuple $\mathcal{(S, A, O, F, R)}$, where $\mathcal{S}$ is the set of environment states, $\mathcal{A}$ is the action space, $\mathcal{O}$ is the observation space, $\mathcal{F: S \times A \to S}$ is the state
transition function, and $\mathcal{R:S\times A \to } \mathbb{R}$ is the reward function. In our setting, which is tailored for LLM agents, the state, action, and observation spaces $\mathcal{(S, A, O)}$ are all represented as natural language sequences over a finite token vocabulary.
% The state $s_t$ is constructed from the most recent $h$ observation-action pairs:
% \begin{equation}
% \label{eq:state}
%     s_t=\{a_{t-h+1},o_{t-h+1}...,a_{t},o_{t}\},
% \end{equation}
% where the history length $h$ may vary depending on the implementation.

At each timestep $t$, the LLM agent $\pi_\theta$ generates an action $a_t$ based on the current state $s_{t-1}$: $a_t \sim \pi_\theta(\cdot|s_{t-1})$. 
After executing the action, the agent receives the environmental feedback as the observation $o_t$. The interaction loop terminates when either the agent completes the task or the maximum step is reached. The final trajectory is $\boldsymbol{\tau} = (\boldsymbol{q},a_1,o_1...,a_n,o_n)$, where $n$ denotes the length of the trajectory and $\boldsymbol{q}$ denotes the task. At termination, a scalar reward $\mathcal{R}(\boldsymbol{\tau})$ is provided. The agent’s objective is to learn an optimal policy $\pi_\theta$ that maximizes the expected cumulative reward:
\begin{equation}
    \max_{\theta}\ \mathbb{E}_{\tau \sim \pi_{\theta}} [\mathcal{R}(\boldsymbol{\tau})].
\end{equation}

\subsection{Reinforcement Learning for Agents}
\textbf{Proximal Policy Optimization (PPO)} PPO~\citep{ppo} is a widely adopted policy gradient algorithm in LLM agent training.
To stabilize training, PPO restricts policy updates to remain within a proximal region of the old policy $\pi_{\boldsymbol{\theta}_{\text{old}}}$ using the following clipped surrogate to maximize the objective:
\begin{align}
\max_{\theta}\ & \mathbb{E}_{\boldsymbol{q} \sim \mathcal{D}, \,
\boldsymbol{\tau} \sim \pi_{\theta_{\text{old}}}(\cdot \mid \boldsymbol{q})}
\Big[ \min\big( r_t(\theta) \hat{A}_t, \nonumber \\
& \quad\; \text{clip}(r_t(\theta), 1 - \epsilon, 1 + \epsilon)\hat{A}_t \big) \Big], 
\label{eq:ppo objective} \\
& \text{with}\quad
r_t(\boldsymbol{\theta}) =
\frac{\pi_{\boldsymbol{\theta}}(\tau_t \mid \boldsymbol{q}, \boldsymbol{\tau}_{<t})}
{\pi_{\boldsymbol{\theta}_{\text{old}}}(\tau_t \mid \boldsymbol{q}, \boldsymbol{\tau}_{<t})}.
\nonumber
\end{align}

Here, $\mathcal{D}$ is a dataset of tasks $\boldsymbol{q}$, $\epsilon \in \mathbb{R}$ is a clip hyperparameter usually set to $0.2$, and $\hat{A}_t$ is the estimated advantage, typically computed via Generalized Advantage Estimation (GAE)~\citep{gae} using a critic network.

Note that while PPO estimates token-level advantages, it falls short compared to our method in several aspects: (1) It relies on a separate critic network, which limits scalability and efficiency. In contrast, our approach introduces only negligible computational overhead and entirely avoids the need for a critic. (2) PPO does not leverage group rollouts or collective reward computation, both of which are integral to our framework and lead to more reliable credit assignment.

\textbf{Group Relative Policy Optimization (GRPO)} 
% Mitigating the memory issue by discarding the critic network and demonstrating stronger performance, group-based RL algorithms~\citep{rloo,yu2025dapo,grpo} have become the mainstream approach recently. One of the most popular one is GRPO~\citep{grpo}.
Building on the clipped objective in \cref{eq:ppo objective}, GRPO discards the critic network by estimating advantages using the average reward within a group of sampled responses. Specifically, for each task $\boldsymbol{q}$, the LLM agent $\pi_{\boldsymbol{\theta}_{\text{old}}}$ generates a group of trajectories $\{\boldsymbol{\tau}_i\}_{i=1}^{G}$ with corresponding outcome rewards $\{R(\boldsymbol{\tau}_i)\}_{i=1}^{G}$, where $G\in \mathbb{R}$ is the group size. The estimated advantage $\hat{A}^i_t$ is then computed as:
% \begin{align}
% \hat{A}^i_t &= \frac{R(\boldsymbol{\tau}_i) - \text{mean}(\{R(\boldsymbol{\tau}_j)\}_{j=1}^{G})}{\text{std}(\{R(\boldsymbol{\tau}_j)\}_{j=1}^{G})}, \\
% &\text{where } R(\boldsymbol{\tau}_i) = \begin{cases}
% 1.0 & \quad \text{if }\ \ \texttt{task\_complete}, \\
% 0.0 & \quad \text{otherwise}.
% \end{cases}
% \label{eq: grpo advantage}
% \end{align}
\begin{align}
\hat{A}^i_t &= \frac{R(\boldsymbol{\tau}_i) - \text{mean}(\{R(\boldsymbol{\tau}_j)\}_{j=1}^{G})}{\text{std}(\{R(\boldsymbol{\tau}_j)\}_{j=1}^{G})} \label{eq: grpo advantage} \\
&\text{where } R(\boldsymbol{\tau}_i) = \begin{cases}
1.0 & \quad \text{if }\ \ \texttt{task\_complete}, \\
0.0 & \quad \text{otherwise}.
\end{cases} \nonumber
\end{align}

In addition to this modified advantage estimation, GRPO adds an explicit KL penalty term to the clipped objective in \cref{eq:ppo objective}.

% \textbf{REINFORCE Leave-One-Out (RLOO)} Compared to GRPO, RLOO~\citep{rloo} constructs a per-sample baseline that is leave-one-out: for every sampled trajectory its baseline is the average return of the other trajectories generated from the same task. RLOO also removes the advantage normalization. The resulting advantage estimate is:
% \begin{equation}
%     \hat{A}^{i}_t=R(\boldsymbol{\tau}_{i})-\text{mean}(\{R(\boldsymbol{\tau}_j)\}_{j=1,j\neq i}^{G}).
% \end{equation}

The group computation used in GRPO and other group-based RL algorithms~\citep{drgrpo,yu2025dapo} is highly memory-efficient and can scale effectively to large batch sizes and model sizes typical in modern LLM training, making it a practical and scalable choice.

\begin{figure*}[t]
  \includegraphics[width=\textwidth]{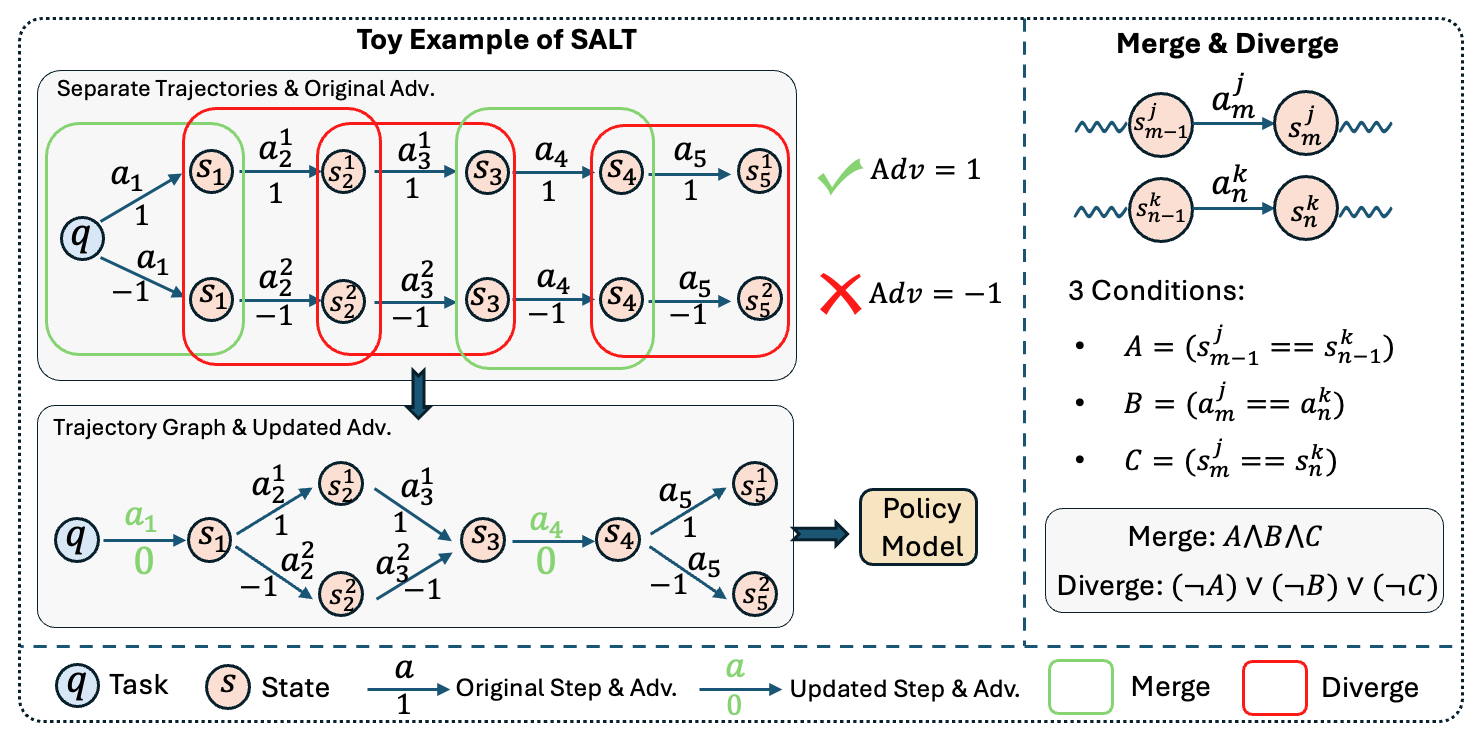}
  \caption{\textbf{Illustration of SALT on a single task with multiple rollouts.} Trajectories from the same prompt are constructed into a trajectory graph using \textit{merge} and \textit{diverge}. Step-level advantages are then refined by averaging advantages over merged edges while preserving original advantages for distinct edges. This yields fine-grained credit assignment using only sparse final rewards.}
  \label{fig:detail}
  \vspace{-0.3cm}
\end{figure*}

\section{Methods}
\label{sec:main}
In this section, we present SALT in detail. Our framework consists of two key components: 
(1) \textbf{Trajectory Graph Construction}, where we build a trajectory graph from multiple rollouts of the same task using \textit{merge} and \textit{diverge} operations; 
(2) \textbf{Step-level Advantage Assignment} where we refine advantages at the step level by leveraging the graph structure and original outcome rewards.

\subsection{Trajectory Graph Construction}
\subsubsection{Graph Initialization}
As introduced in Preliminary~\ref{sec:pre_agents}, given a task $\boldsymbol{q}$, group-based RL algorithms generate a set of trajectories $\{\boldsymbol{\tau}_i\}_{i=1}^{G}$, each associated with a scalar outcome reward $R(\boldsymbol{\tau}_i)$. These rewards are used to compute group-normalized advantages $\{\hat{A}^{i}\}_{i=1}^{G}$ (Eq.~\ref{eq: grpo advantage}). Within each trajectory $\boldsymbol{\tau}_i = (\boldsymbol{q}, a_1^i, o_1^i, \dots, a_{n_i}^i, o_{n_i}^i)$, all steps $(a_1^i, \dots, a_{n_i}^i)$ are assigned the same advantage $\hat{A}^{i}$ — meaning that regardless of their individual quality, all actions are uniformly rewarded or penalized.

However, as we mentioned, we observe that beneficial and detrimental actions are often entangled across multi-step interactions. To address this, SALT replaces trajectory-level advantages with fine-grained, step-level advantages $\{\hat{A}_s^{i}\}_{i=1}^{G}$.

To achieve this, we construct a directed acyclic trajectory graph $\mathcal{G} = (V, E, H)$ over the set of trajectories $\{\boldsymbol{\tau}_i\}_{i=1}^{G}$ for task $\boldsymbol{q}$.
\begin{itemize}[leftmargin=*]
\item $V$ (nodes) represents all states, including:
\begin{itemize}
    \item The task description $\boldsymbol{q}$ as the root node, and
    \item All subsequent states across trajectories:
\end{itemize}
\begin{equation}
    V = \{\boldsymbol{q},\underbrace{s_1^1,s_2^1,\cdots s_{n_1}^1}_{traj.\ 1},\cdots , \underbrace{s_1^G,s_2^G,...s_{n_G}^G}_{traj.\ G}\}.
\end{equation}

\item $E$ (edges) represents all actions in the form of tuples:
% \begin{equation}
% \label{eq:act}
%     E=\{\underbrace{(\boldsymbol{q},a_1^1,s_1^1),(\boldsymbol{q},a_2^1,s_2^1),\cdots (\boldsymbol{q},a_{n_1}^1,s_{n_1}^1)}_{traj.\ 1}, \cdots , \underbrace{(\boldsymbol{q},a_1^G,s_1^G),(\boldsymbol{q},a_2^G,s_2^G),...(\boldsymbol{q},a_{n_G}^G,s_{n_G}^G)}_{traj.\ G}\}.    
% \end{equation}
\begin{equation}
\label{eq:act}
\begin{split}
\hspace{-0.2cm}E=\{&\underbrace{(\boldsymbol{q},a_1^1,s_1^1),(s_1^1,a_2^1,s_2^1),\cdots (s_{n_1-1}^1,a_{n_1}^1,s_{n_1}^1)}_{traj.\ 1}, \\
&\qquad\qquad\qquad\qquad\cdots, \\
&\hspace{-1cm}\underbrace{(\boldsymbol{q},a_1^G,s_1^G),(s_1^G,a_2^G,s_2^G),...(s_{n_G-1}^G,a_{n_G}^G,s_{n_G}^G)}_{traj.\ G}\}.
\end{split}
\end{equation}

\item $H$ (advantage values) stores the initial advantage for each action:
\begin{equation}
    H=\{\underbrace{\hat{A}_1^1,\hat{A}_2^1,\cdots \hat{A}_{n_1}^1}_{traj.\ 1}, \cdots , \underbrace{\hat{A}_1^G,\hat{A}_2^G,...\hat{A}_{n_G}^G}_{traj.\ G}\}.
\end{equation}
\end{itemize}

% Specifically, $V$ represents the node set of states consisting of 1. the task description $\boldsymbol{q}$ as the start node and 2. all other states:
% \begin{equation}
%     V = \{\boldsymbol{q},\underbrace{s_1^1,s_2^1,\cdots s_{n_1}^1}_{traj.\ 1},\cdots , \underbrace{s_1^G,s_2^G,...s_{n_G}^G}_{traj.\ G}\}.
% \end{equation}

% Similarly, $E$ denotes the edge set consisting of all the actions:
% \begin{equation}
% \label{eq:act}
    % E=\{\underbrace{a_1^1,a_2^1,\cdots a_{n_1}^1}_{traj.\ 1}, \cdots , \underbrace{a_1^G,a_2^G,...a_{n_G}^G}_{traj.\ G}\}.    
% \end{equation}
% Each edge $a_i^j$ has the corresponding advantage value $\hat{A}_i^j$ which constitutes $H$:
% \begin{equation}
%     H=\{\underbrace{\hat{A}_1^1,\hat{A}_2^1,\cdots \hat{A}_{n_1}^1}_{traj.\ 1}, \cdots , \underbrace{\hat{A}_1^G,\hat{A}_2^G,...\hat{A}_{n_G}^G}_{traj.\ G}\}.    
% \end{equation}

Initially, the graph shares a common root node $\boldsymbol{q}$, with $G$ branches diverging from it. Within each branch, all edges inherit the same trajectory-level advantage (i.e., $\hat{A}_1^i = \hat{A}_2^i = \cdots = \hat{A}_{n_i}^i$).

\subsubsection{Graph Refinement}
We then refine the graph using two operations — \textit{merge} and \textit{diverge}:

\paragraph{Merge Operation.}
Two edges $(s_{m-1}^j, a_m^j, s_{m}^j)$ and $(s_{n-1}^k, a_n^k, s_{n}^k)$ are merged if \textbf{all} of the following hold:
\begin{itemize}
    \item $s_{m-1}^j = s_{n-1}^k$: same starting state,
    \item $a_m^j = a_n^k$: same action taken,
    \item $s_{m}^j = s_{n}^k$: same resulting state.
\end{itemize}

\paragraph{Diverge Operation.}
Two edges diverge if \textbf{any one} of the following is true:
\begin{itemize}
    \item $s_{m-1}^j \neq s_{n-1}^k$: different starting states,
    \item $s_{m-1}^j = s_{n-1}^k$ but $a_m^j \neq a_n^k$: same state, different action,
    \item $s_{m-1}^j = s_{n-1}^k$, $a_m^j = a_n^k$, but $s_{m}^j \neq s_{n}^k$: same state and action, but different resulting state.
\end{itemize}

% \textbf{Merge} Consider two edges $(s_{m-1}^j,a_m^j,s_{m}^j)$ and $(s_{n-1}^k,a_n^k,s_{n}^k)$, they merge when it meets \textbf{all of} the following criteria:
% \begin{itemize}
%     \item $s_{m-1}^j=s_{n-1}^k$: Agent is under the same state, \textbf{AND}
%     \item $a_m^j=a_n^k$: Agent takes the same action under the same state, \textbf{AND}
%     \item $s_{m}^j=s_{n}^k$: Agent reaches the same state after taking the same action.
% \end{itemize}

% \textbf{Diverge} Consider two edges $(s_{m-1}^j,a_m^j,s_{m}^j)$ and $(s_{n-1}^k,a_n^k,s_{n}^k)$, they diverge when it meets \textbf{one of} the three criteria:
% \begin{itemize}
%     \item $s_{m-1}^j\neq s_{n-1}^k$: Agent is under different state, \textbf{OR}
%     \item $s_{m-1}^j=s_{n-1}^k, a_m^j\neq a_n^k$: Agent is under the same state, but takes different action, \textbf{OR}
%     \item $s_{m-1}^j=s_{n-1}^k,a_m^j=a_n^k, s_{m}^j\neq s_{n}^k$: Agent is under the same state and takes the same action, but reaches different states.
% \end{itemize}

% Note that these two operations can only happen across different trajectories instead of within one trajectory ($j\neq k$). 
Note that we define the state $s_t$ as the sequence of the most recent $h$ observation-action pairs, i.e., $s_t=\{a_{t-h+1},o_{t-h+1}...,a_{t},o_{t}\}$, rather than relying solely on the immediate observation $o_t$. This design enables SALT to capture richer contextual dependencies across steps. The history length $h$ serves as a tunable hyperparameter that controls the strictness of merge and diverge operations: larger $h$ encourages stricter state matching, while smaller $h$ allows more flexible matching.

We construct the trajectory graph $\mathcal{G}' = (V', E', H)$ in a greedy, sequential manner: for each trajectory in the group, we process its steps from start to end, incrementally integrating them into the evolving graph via \textit{merge} and \textit{diverge} operations. This online construction yields a refined graph that explicitly encodes which steps are shared across trajectories and which are distinct.

% After iteratively applying these two operations on all the rollouts in a group one by one, we obtain a refined graph $\mathcal{G}' = (V', E', H)$, whose structure explicitly encodes which steps are shared across trajectories and which are distinct.

\subsection{Step-level Advantage Assignment}
Given the refined graph $\mathcal{G}'$, we now reassign advantages to better reflect each step’s contribution.

Suppose there are $M$ sets of merged edges: $\{E_1, E_2, \dots, E_M \mid E_i \subset E\}$, where each $E_i = \{a_1, a_2, \dots, a_{n_i}\}$ and corresponding advantages $H_i = \{\hat{A}_1, \hat{A}_2, \dots, \hat{A}_{n_i}\}$ (we omit superscripts here for clarity). For each merged set, we update all associated advantages to their group mean:
\begin{equation}
    \hat{A}_1' = \hat{A}_2' = \cdots = \hat{A}_{n_i}' 
    = \frac{1}{n_i} \sum_{j=1}^{n_i} \hat{A}_j,
\end{equation}
where $\hat{A}'$ denotes the updated advantages.

The intuition is simple: steps that appear identically across multiple trajectories are likely ``neutral'' or ``required'' — they should not be over-penalized or over-rewarded based on any single trajectory’s outcome. Averaging their advantages reduces gradient conflict and stabilizes training.

For divergent edges, those unique to specific trajectories, we retain their original advantages. These steps are the true differentiators: they directly influence whether a trajectory succeeds or fails, and thus deserve trajectory-specific credit.

The result is an updated advantage set $H'$, containing fine-grained, step-level signals ready to guide policy updates as in \cref{eq:ppo objective}

\begin{table*}[t]
\centering
\resizebox{\textwidth}{!}{
\begin{tabular}{ll|ccccccc|cc}
\toprule
\multirow{2}{*}{\textbf{Type}} & \multirow{2}{*}{\textbf{Method}} & \multicolumn{7}{c|}{\textbf{ALFWorld}} & \multicolumn{2}{c}{\textbf{WebShop}} \\
 & & Pick & Look & Clean & Heat & Cool & Pick2 & All & Score & Succ.\\
\midrule
\rowcolor{gray!15}\multicolumn{11}{c}{\textit{Base: Closed-Source Models}} \\
\multirow{2}{*}{Prompting}& GPT-4o & 75.3 & 60.8 & 31.2 & 56.7 & 21.6 & 49.8 & 48.0& 31.8 & 23.7\\
& Gemini-2.5-Pro & 92.8 & 63.3 & 62.1 & 69.0 & 26.6 & 58.7 & 60.3& 42.5 & 35.9\\
\midrule
\rowcolor{gray!15}\multicolumn{11}{c}{\textit{Base: Qwen2.5-1.5B-Instruct}} \\
\multirow{3}{*}{Prompting}& Qwen2.5 & 5.9 & 5.5 & 3.3 & 9.7 & 4.2 & 0.0 & 4.1 & 23.1 & 5.2\\
& ReAct & 17.4 & 20.5 & 15.7 & 6.2 & 7.7 & 2.0 & 12.8& 40.1& 11.3\\
& Reflexion & 35.3 & 22.2 & 21.7 & 13.6 & 19.4 & 3.7 & 21.8 & 55.8& 21.9\\
\cmidrule{1-11}
\multirow{5}{*}{RL Training}& PPO & 93.7\textsubscript{\textpm1.1} & 80.0\textsubscript{\textpm9.4} & 93.6\textsubscript{\textpm3.5} & 83.1\textsubscript{\textpm6.0} & 74.6\textsubscript{\textpm7.7} & 70.8\textsubscript{\textpm6.8} & 83.5\textsubscript{\textpm1.4}& 85.8\textsubscript{\textpm2.1} & 72.6\textsubscript{\textpm3.2} \\
\cmidrule{2-11}
& RLOO & 91.3\textsubscript{\textpm2.9} & \textbf{77.0}\textsubscript{\textpm15.5} & 79.9\textsubscript{\textpm7.5} & 79.2\textsubscript{\textpm1.7} & 73.2\textsubscript{\textpm5.5} & \textbf{68.8}\textsubscript{\textpm2.5} & 79.2\textsubscript{\textpm3.5}& 84.1\textsubscript{\textpm1.4}& 71.6\textsubscript{\textpm2.4}\\
& \cellcolor{blue!10}RLOO+SALT & \cellcolor{blue!10}\textbf{94.2}\textsubscript{\textpm2.1} & \cellcolor{blue!10}74.4\textsubscript{\textpm9.9} & \cellcolor{blue!10}\textbf{95.0}\textsubscript{\textpm1.4} & \cellcolor{blue!10}\textbf{89.5}\textsubscript{\textpm7.5} & \cellcolor{blue!10}\textbf{79.5}\textsubscript{\textpm8.6} & \cellcolor{blue!10}64.9\textsubscript{\textpm4.6} & \cellcolor{blue!10}\textbf{84.0}\textsubscript{\textpm2.2}& \cellcolor{blue!10}\textbf{87.9}\textsubscript{\textpm0.8}& \cellcolor{blue!10}\textbf{75.8}\textsubscript{\textpm2.9}\\
\cmidrule{2-11}
& GRPO & 92.1\textsubscript{\textpm1.6} & 64.3\textsubscript{\textpm20.2} & 89.1\textsubscript{\textpm5.7} & \textbf{84.1}\textsubscript{\textpm4.8} & 75.3\textsubscript{\textpm7.8} & 75.3\textsubscript{\textpm9.0} & 81.8\textsubscript{\textpm2.1}& 86.2\textsubscript{\textpm2.1} & 72.2\textsubscript{\textpm4.1}\\
& \cellcolor{blue!10}GRPO+SALT & \cellcolor{blue!10}\textbf{96.2}\textsubscript{\textpm1.7} & \cellcolor{blue!10}\textbf{65.2}\textsubscript{\textpm10.8} & \cellcolor{blue!10}\textbf{93.1}\textsubscript{\textpm4.7} & \cellcolor{blue!10}81.8\textsubscript{\textpm8.3} & \cellcolor{blue!10}\textbf{85.0}\textsubscript{\textpm6.9} & \cellcolor{blue!10}\textbf{77.0}\textsubscript{\textpm4.7} & \cellcolor{blue!10}\textbf{85.2}\textsubscript{\textpm2.5}& \cellcolor{blue!10}\textbf{86.9}\textsubscript{\textpm0.6}& \cellcolor{blue!10}\textbf{74.7}\textsubscript{\textpm2.4}\\
\midrule
\rowcolor{gray!15}\multicolumn{11}{c}{\textit{Base: Qwen2.5-7B-Instruct}} \\
\multirow{3}{*}{Prompting}& Qwen2.5 & 33.4 & 21.6 & 19.3 & 6.9 & 2.8 & 3.2 & 14.8 & 26.4 & 7.8\\
& ReAct & 48.5 & 35.4 & 34.3 & 13.2 & 18.2 & 17.6 & 31.2 & 46.2 & 19.5\\
& Reflexion & 62.0 & 41.6 & 44.9 & 30.9 & 36.3 & 23.8 & 42.7& 58.1& 28.8\\
\cmidrule{1-11}
\multirow{5}{*}{RL Training}& PPO & 97.0\textsubscript{\textpm1.2} & 66.3\textsubscript{\textpm6.7} & 93.2\textsubscript{\textpm2.3} & 95.7\textsubscript{\textpm3.1} & 72.4\textsubscript{\textpm4.0} & 75.3\textsubscript{\textpm1.8} 
& 85.5\textsubscript{\textpm1.3} & 77.9\textsubscript{\textpm4.3}& 71.5\textsubscript{\textpm5.0}\\
\cmidrule{2-11}
& RLOO & 93.1\textsubscript{\textpm1.6} & 70.6\textsubscript{\textpm6.9} & 78.3\textsubscript{\textpm3.8} & 86.9\textsubscript{\textpm7.0} & 69.7\textsubscript{\textpm3.9} & 67.5\textsubscript{\textpm7.9} & 79.3\textsubscript{\textpm0.7} & \textbf{83.6}\textsubscript{\textpm2.6} & \textbf{76.8}\textsubscript{\textpm3.6}\\
& \cellcolor{blue!10}RLOO+SALT & \cellcolor{blue!10}\textbf{93.4}\textsubscript{\textpm2.1} & \cellcolor{blue!10}\textbf{72.6}\textsubscript{\textpm7.5} & \cellcolor{blue!10}\textbf{91.5}\textsubscript{\textpm3.2} & \cellcolor{blue!10}\textbf{90.3}\textsubscript{\textpm3.5} & \cellcolor{blue!10}\textbf{78.1}\textsubscript{\textpm2.7} & \cellcolor{blue!10}\textbf{76.4}\textsubscript{\textpm4.4} & \cellcolor{blue!10}\textbf{87.3}\textsubscript{\textpm4.4}& \cellcolor{blue!10}83.1\textsubscript{\textpm3.8}& \cellcolor{blue!10}75.2\textsubscript{\textpm5.5}\\
\cmidrule{2-11}
& GRPO & \textbf{88.9}\textsubscript{\textpm3.9} & \textbf{67.0}\textsubscript{\textpm8.7} & 73.7\textsubscript{\textpm11.7} & \textbf{76.1}\textsubscript{\textpm7.7} & 52.7\textsubscript{\textpm8.1} & 68.2\textsubscript{\textpm6.4} & 72.5\textsubscript{\textpm5.5} & 79.8\textsubscript{\textpm4.0} & 72.4\textsubscript{\textpm5.5}\\
& \cellcolor{blue!10}GRPO+SALT & \cellcolor{blue!10}87.5\textsubscript{\textpm4.9} & \cellcolor{blue!10}58.8\textsubscript{\textpm14.7} & \cellcolor{blue!10}\textbf{89.3}\textsubscript{\textpm3.6} & \cellcolor{blue!10}75.3\textsubscript{\textpm4.3} & \cellcolor{blue!10}\textbf{70.2}\textsubscript{\textpm5.5} & \cellcolor{blue!10}\textbf{68.5}\textsubscript{\textpm5.8} & \cellcolor{blue!10}\textbf{77.8}\textsubscript{\textpm1.7}& \cellcolor{blue!10}\textbf{84.7}\textsubscript{\textpm1.7}& \cellcolor{blue!10}\textbf{76.2}\textsubscript{\textpm3.4}\\
\bottomrule
\end{tabular}
}
\caption{\textbf{Performance on ALFWorld and WebShop}. The results are reported as the average and standard deviation of three independent trainings.}
\label{tab:web_main}
\vspace{-0.3cm}
\end{table*}

\section{Experiments}
\subsection{Experimental Setup}
\textbf{Benchmarks.} We evaluate our method on three challenging long-horizon agent benchmarks:  
(1) \textbf{WebShop}~\citep{webshop}: a simulated e-commerce environment with real products and crowd-sourced instructions, where agents must navigate search, results, and product pages, performing actions like querying, filtering, and selecting, to complete purchase tasks.  
(2) \textbf{ALFWorld}~\citep{ALFWorld20}: a platform that bridges abstract text-based environments with embodied tasks from ALFRED, requiring agents to reason before executing physical actions.  
(3) \textbf{AppWorld}~\citep{appworld}: a suite of nine simulated consumer apps (e.g., email, payments, music, shopping, phone, file system), testing agents’ ability to invoke complex APIs to fulfill user requests.
%(4) Sokoban~\citep{SchraderSokoban2018} which is a warehouse keeper puzzle game where agents must push boxes onto storage locations/targets. It features randomly generated rooms with five main elements: walls, floor, boxes, box targets, and a player, along with nine possible actions including pushing and moving in different directions. 

\textbf{Baselines.}  
We compare SALT against three categories of strong baselines: (1) \textit{prompting-based} (training-free) methods, (2) \textit{supervised fine-tuning} (SFT) approaches, and (3) \textit{reinforcement learning} (RL) algorithms. To evaluate the effectiveness of our step-level advantage assignment, we integrate SALT into two representative group-based RL methods, GRPO~\citep{grpo} and RLOO~\citep{rloo}, and measure the performance gain it brings as a plug-and-play enhancement.

% We compare SALT against two categories of strong baselines:  
% (1) \textit{Prompting-based (training-free)}: closed-source models including GPT-4o~\citep{gpt4o}, and Gemini 2.5 Pro~\citep{comanici2025gemini}; and open-source models Qwen2.5~\citep{qwen2.5}, prompted in ReAct~\citep{yao2023react} or Reflexion~\citep{shinn2023reflexion} style.  
% (2) \textit{RL-based}: PPO~\citep{ppo} (with critic), and two group-based algorithms — RLOO~\citep{rloo} and GRPO~\citep{grpo} — which serve as our primary baselines for integration with SALT.
% Results for prompting-based methods are adopted from~\citet{gigpo} and~\citet{loop}. Performance of PPO, RLOO, and GRPO is evaluated using our own implementations to ensure fair and consistent comparisons under identical training conditions.

\textbf{Training Setup.}  
We use Qwen2.5-32B-Instruct for AppWorld, Qwen2.5-1.5B-Instruct/Qwen2.5-7B-Instruct model for WebShop and ALFWorld, and set the state history length to 3 when constructing the trajectory graph.
The group size for all datasets is set to 8.
For AppWorld, we employ the BGE-M3 model~\citep{chen2024bge} as the embedder to encode both states and steps; embeddings with a cosine similarity greater than 0.8 are considered equivalent.
All group-based RL methods (including SALT variants) share identical hyperparameters to ensure fair comparison.

\textbf{Evaluation Setup.}  
For \textit{AppWorld}, we report Task Goal Completion (TGC) and Scenario Goal Completion (SGC), success rates per task and per scenario, on both test-normal (Test-N) and test-challenge (Test-C) splits. Results are averaged over three independent runs, and we report both the mean and standard deviation.
For \textit{ALFWorld} and \textit{WebShop}, we report metrics averaged over the last five checkpoints to account for high variance: for \textit{ALFWorld}, average success rates per subtask and overall; for \textit{WebShop}, both average normalized score and success rate.

Full experimental setup and details are provided in Appendix~\ref{app:exp}.

\subsection{Overall Results}
Summarized in Tables~\ref{tab:web_main} and~\ref{tab:app_main}, our results show that SALT consistently enhances group-based RL algorithms, though its effectiveness depends on model scale and task structure.

On ALFWorld, SALT improves both GRPO and RLOO with the Qwen2.5-1.5B model, boosting GRPO from $81.8\% \rightarrow 85.2\%$ (+3.4pp) and RLOO from $79.2\% \rightarrow 84.0\%$ (+4.8pp), with large subtask gains such as \textit{Cool} (+9.7pp) and \textit{Heat} (+10.3pp). With the 7B model, SALT continues to help: RLOO improves from $79.3\% \rightarrow 87.3\%$ (+8.0pp), and GRPO from $72.5\% \rightarrow 77.8\%$ (+5.3pp). On WebShop, SALT benefits the 1.5B model (RLOO: $71.6\% \rightarrow 75.8\%$, +4.2pp), but on the 7B model, RLOO slightly declines ($76.8\% \rightarrow 75.2\%$, –1.6pp) while GRPO still gains (+3.8pp), suggesting that SALT’s credit assignment may interact differently with larger-model optimization dynamics.

The gains are more pronounced on the complex AppWorld benchmark. SALT consistently lifts RLOO, e.g., on Test-C, TGC improves from $33.8\% \rightarrow 37.4\%$ (+3.6pp) and SGC from $14.4\% \rightarrow 18.9\%$ (+4.5pp). GRPO+SALT achieves the best overall performance, with TGC on Test-N rising from $61.5\% \rightarrow 66.2\%$ (+4.7pp) and on Test-C from $30.1\% \rightarrow 36.8\%$ (+6.7pp), demonstrating SALT’s ability to enhance generalization in challenging, novel tasks.

Notably, SALT outperforms PPO, despite PPO using a critic, validating that fine-grained credit assignment from outcome rewards alone is both feasible and highly effective. Together, these results confirm SALT’s robustness across model sizes and task complexities.

\begin{table}[t]
\centering
\resizebox{\columnwidth}{!}{
\begin{tabular}{ll|cc|cc}
\toprule
 \multirow{2}{*}{\textbf{Type}}& \multirow{2}{*}{\textbf{Method}}& \multicolumn{2}{c|}{\textbf{Test-N}} & \multicolumn{2}{c}{\textbf{Test-C}} \\ 
&  & TGC & SGC & TGC & SGC \\
\midrule
\rowcolor{gray!15}\multicolumn{6}{c}{\textit{Base: Closed-Source Models}} \\
\multirow{4}{*}{Prompting}& GPT-4o & 48.8 & 32.1 & 30.2 & 13 \\
 & OpenAI o1 & 61.9 & 41.1 & 36.7 & 19.4\\
 & Llama 3 70B & 24.4 & 17.9 & 7.0 & 4.3 \\
 & Qwen 2.5 32B & 39.2\textsubscript{\textpm3.5} & 18.6\textsubscript{\textpm2.0} & 
21.0\textsubscript{\textpm1.4} & 7.5\textsubscript{\textpm1.2} \\ 
\midrule
\rowcolor{gray!15}\multicolumn{6}{c}{\textit{Base: Qwen2.5-32B-Instruct}} \\
 \multirow{3}{*}{SFT} & SFT-GT & 6.2\textsubscript{\textpm0.7} & 1.8\textsubscript{\textpm0.0} & 0.8\textsubscript{\textpm0.2} & 0.1\textsubscript{\textpm0.3}\\ 
  & RFT & 47.9\textsubscript{\textpm3.7} & 26.4\textsubscript{\textpm2.3} & 26.4\textsubscript{\textpm1.8} & 11.4\textsubscript{\textpm2.3} \\ 
  & EI & 58.3\textsubscript{\textpm2.8} & 36.8\textsubscript{\textpm6.0} & 32.8\textsubscript{\textpm0.7} & 17.6\textsubscript{\textpm1.3} \\ 
 \midrule 
\multirow{7}{*}{RL Training} & DPO-MCTS & 57.0\textsubscript{\textpm1.5} & 31.8\textsubscript{\textpm4.2} & 31.8\textsubscript{\textpm1.3} & 13.7\textsubscript{\textpm1.5} \\ 
  & DMPO & 59.0\textsubscript{\textpm1.2} & 36.6\textsubscript{\textpm4.7} & 36.3\textsubscript{\textpm1.8} & 18.4\textsubscript{\textpm2.3} \\ 
  & PPO & 50.8\textsubscript{\textpm3.7} & 28.9\textsubscript{\textpm7.9} & 26.4\textsubscript{\textpm0.5} & 10.5\textsubscript{\textpm2.1} \\ 
  \cmidrule{2-6}
  & RLOO & 59.8\textsubscript{\textpm2.2} & 37.5\textsubscript{\textpm2.8} & 33.8\textsubscript{\textpm0.5} & 14.4\textsubscript{\textpm1.6} \\ 
& \cellcolor{blue!10}RLOO+SALT & \cellcolor{blue!10}\textbf{61.3}\textsubscript{\textpm2.7} & \cellcolor{blue!10}\textbf{39.3}\textsubscript{\textpm4.1} & \cellcolor{blue!10}\textbf{37.4}\textsubscript{\textpm0.9} & \cellcolor{blue!10}\textbf{18.9}\textsubscript{\textpm1.2} \\
\cmidrule{2-6}
& GRPO & 61.5\textsubscript{\textpm2.9} & 41.4\textsubscript{\textpm3.8} & 36.3\textsubscript{\textpm0.2} & 17.0\textsubscript{\textpm0.9} \\ 
& \cellcolor{blue!10}GRPO+SALT & \cellcolor{blue!10}\textbf{66.2}\textsubscript{\textpm2.5} & \cellcolor{blue!10}\textbf{47.9}\textsubscript{\textpm4.1} & \cellcolor{blue!10}\textbf{36.8}\textsubscript{\textpm1.5} & \cellcolor{blue!10}\textbf{20.9}\textsubscript{\textpm1.8} \\
\bottomrule
\end{tabular}}
\caption{\textbf{Performance on AppWorld}. The results are reported as the average and standard deviation of three independent evaluations.}
\label{tab:app_main}
\vspace{-0.3cm}
\end{table}

% \begin{table*}[t]
% \centering
% \label{tab:main}
% \resizebox{\columnwidth}{!}{
% \begin{tabular}{l|ccc|ccc|ccc|ccc}
% \toprule
% \multirow{2}{*}{\textbf{Method}} & \multicolumn{3}{c|}{\textbf{Test-N TGC}} & \multicolumn{3}{c|}{\textbf{Test-N SGC}} & \multicolumn{3}{c|}{\textbf{Test-C TGC}} & \multicolumn{3}{c}{\textbf{Test-C SGC}}\\
% % \cmidrule(r){2-4}\cmidrule(r){5-7}\cmidrule(r){8-10}\cmidrule(r){11-13}
% & D1 & D2 & D3 & D1 & D2 & D3 & D1 & D2 & D3 & D1 & D2 & D3 \\
% \midrule
% RLOO & 0\textsubscript{\textpm0} & 0\textsubscript{\textpm0} & 0\textsubscript{\textpm0} & 0\textsubscript{\textpm0} & 0\textsubscript{\textpm0} & 0\textsubscript{\textpm0} & 0\textsubscript{\textpm0} & 0\textsubscript{\textpm0} & 0\textsubscript{\textpm0} & 0\textsubscript{\textpm0} & 0\textsubscript{\textpm0} & 0\textsubscript{\textpm0} \\
% RLOO+SALT& 0\textsubscript{\textpm0} & 0\textsubscript{\textpm0} & 0\textsubscript{\textpm0} & 0\textsubscript{\textpm0} & 0\textsubscript{\textpm0} & 0\textsubscript{\textpm0} & 0\textsubscript{\textpm0} & 0\textsubscript{\textpm0} & 0\textsubscript{\textpm0} & 0\textsubscript{\textpm0} & 0\textsubscript{\textpm0} & 0\textsubscript{\textpm0} \\
% \bottomrule
% \end{tabular}}
% \caption{Performance on AppWorld.}
% \end{table*}

\subsection{Compute Efficiency}
\begin{figure}[h]
  \includegraphics[width=\columnwidth]{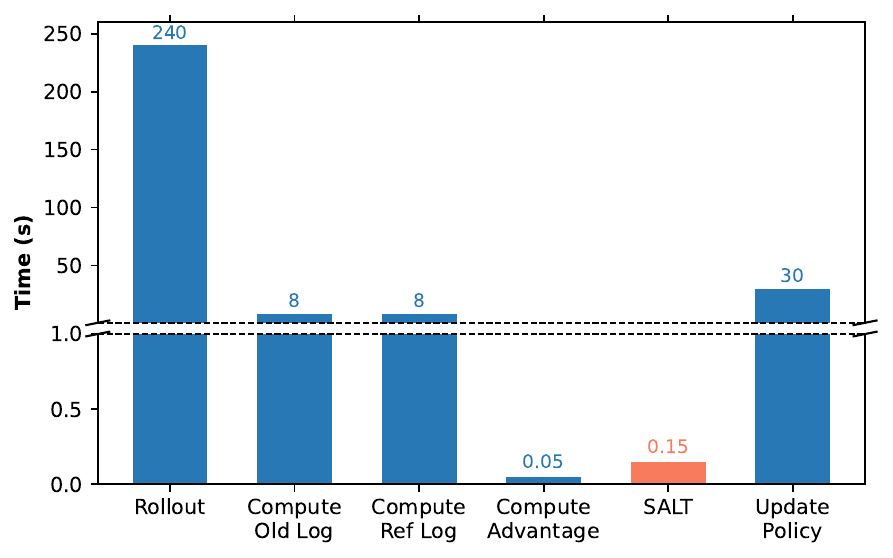}
  \caption{Time costs for different components per step.}
  \label{fig:time_cost}
\end{figure}

SALT is designed as a lightweight, plug-and-play module that integrates seamlessly into existing group-based RL pipelines — inserted after advantage computation and before policy update. As shown in Figure~\ref{fig:time_cost}, we break down the computational cost of each component during training on ALFWorld using GRPO with Qwen2.5-1.5B-Instruct.
The dominant time-consuming operations are rollout (240s) and policy update (30s), while computing old and reference log probabilities each takes about 8s. In contrast, the original advantage computation requires only 0.05s. Remarkably, SALT's step-level advantage generation (SAG) adds just 0.15s — a negligible increase over the baseline, and less than three times the cost of standard advantage estimation.
This demonstrates that SALT introduces minimal computational overhead, making it highly efficient and scalable for large-scale agent training.

\subsection{Impact of Model Size}
\begin{figure}[h]
  \includegraphics[width=\columnwidth]{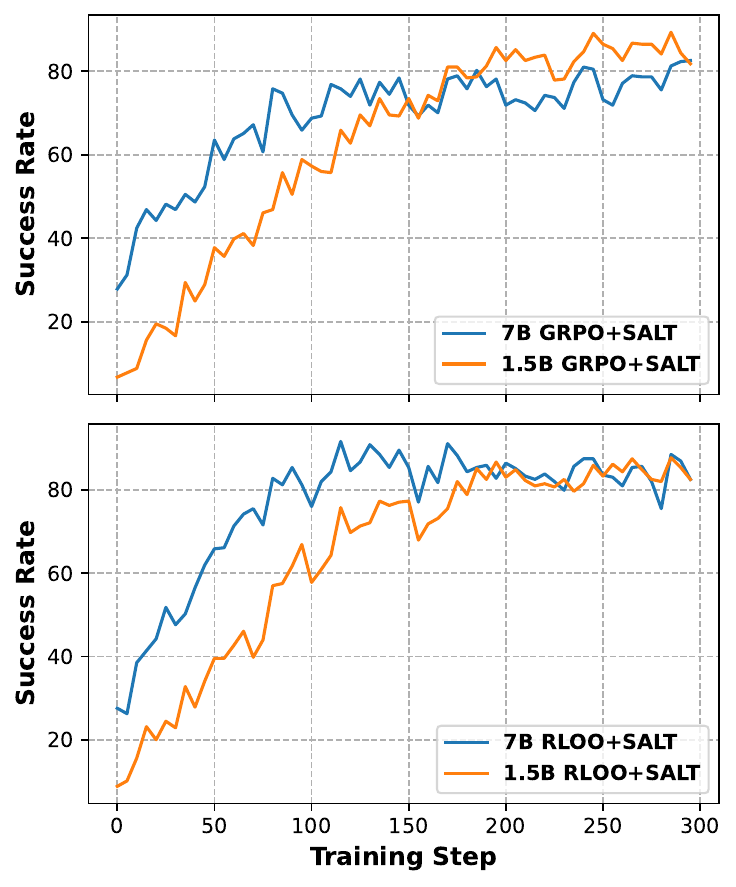}
  \vspace{-0.8cm}
  \caption{Success rate over training steps for 1.5B and 7B models.}
  \label{fig:model_size}
  \vspace{-0.3cm}
\end{figure}

We examine how model size affects performance by comparing Qwen2.5-1.5B-Instruct and Qwen2.5-7B-Instruct on ALFWorld, as shown in Figure~\ref{fig:model_size}. Counterintuitively, the smaller 1.5B model eventually matches or exceeds the 7B variant, despite initial lag.
We attribute this phenomena to the \textit{exploration-exploitation dynamics}.
First, smaller models exhibit a more pronounced \textit{exploration bias} during early training. Due to their limited capacity, they are less able to precisely fit observed reward patterns, which counterintuitively encourages broader exploration of the action space. This often leads them to discover high-reward trajectories that larger models overlook. As training progresses and SALT provides increasingly accurate step-level advantages, these exploratory gains are rapidly consolidated into stable, high-performing policies.
In contrast, larger models demonstrate stronger \textit{exploitation capabilities} early on: they quickly latch onto rewarding action sequences and refine them. However, this strength can become a liability as their high capacity and precise fitting may cause them to converge prematurely to local optima. 

\subsection{Impact of State History Length}
\begin{figure}[h]
  \includegraphics[width=\columnwidth]{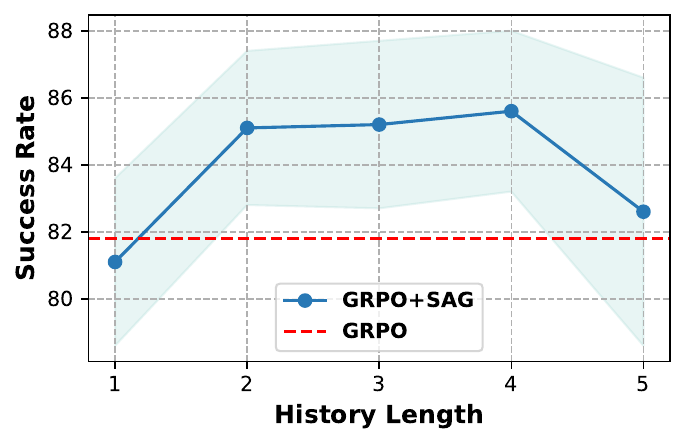}
  \vspace{-0.8cm}
  \caption{Success rate with varying state history lengths.}
  \label{fig:success_history}
  \vspace{-0.3cm}
\end{figure}

The state history length $h$ is a key hyperparameter in SALT, determining the granularity of merge and diverge operations. We have defined the state as $s_t = (a_{t-h+1}, o_{t-h+1}, \dots, a_t, o_t)$, where $h$ controls how much historical context is considered.
We evaluate its impact on ALFWorld using GRPO with Qwen2.5-1.5B-Instruct, varying $h \in \{1,2,3,4,5\}$. As shown in Figure~\ref{fig:success_history}, when $h=1$, SALT underperforms vanilla GRPO due to excessive merging: many steps are incorrectly grouped, introducing noise into advantage signals. As $h$ increases to 2 or 3, performance improves significantly, indicating an optimal balance between merging and divergence. However, at $h=5$, performance drops back toward baseline levels, as the strict matching criteria reduce merge frequency, causing SALT to degenerate into trajectory-level updates.

\begin{figure}[h]
  \includegraphics[width=\columnwidth]{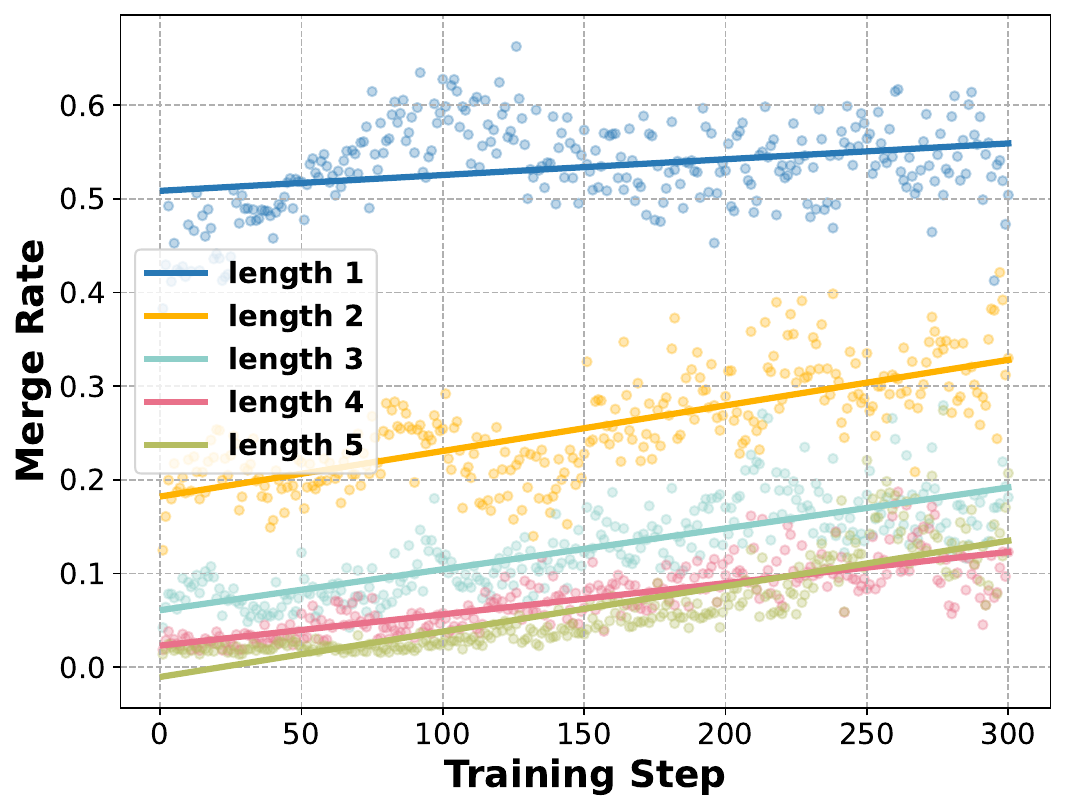}
  \vspace{-0.8cm}
  \caption{Merge rate across training steps for different state history lengths.}
  \label{fig:merge_rate}
  \vspace{-0.3cm}
\end{figure}

This trend aligns with the merge rate dynamics shown in Figure~\ref{fig:merge_rate}. The \textit{merge rate} is defined as the proportion of merge operations to the total number of merge and diverge operations during trajectory graph construction. 
Higher $h$ leads to lower merge rates, as more stringent state alignment is required for merging, making it harder for steps across trajectories to be considered equivalent. 
And over training, the merge rate increases for all $h$ values, reflecting the model’s growing determinism and convergence in behavior. This suggests that SALT adapts naturally to the model’s evolving policy, becoming more effective as training progresses.

% \subsection{Dynamics of SALT}
% As the model undergoes training, we examine the dynamics of SALT by computing the Merge Rate, which is defined as the number of merge operations over all operations. 

% We also show the curve of success rate during test time and the average reward during training test in~\cref{}. Clearly, integrating SALT consistently improves the performance compared to the vanilla GRPO algorithm.

\subsection{Impact of Group Size}
\begin{figure}[h]
  \includegraphics[width=\columnwidth]{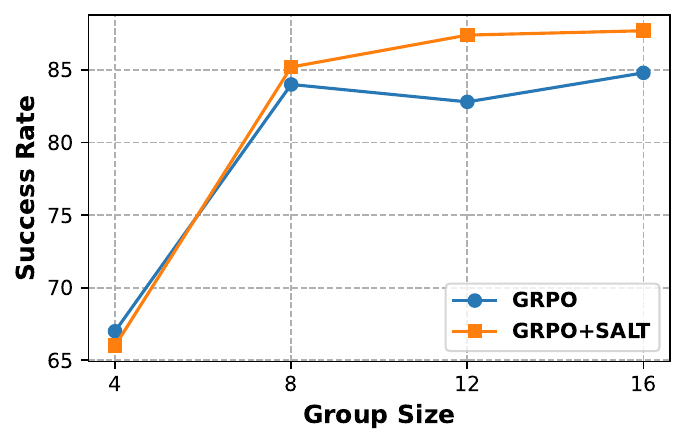}
  \vspace{-0.8cm}
  \caption{Success rate across different group sizes.}
  \label{fig:group_size}
  \vspace{-0.3cm}
\end{figure}

Group size is a critical hyperparameter in group-based RL, balancing training stability against computational cost. Larger groups provide more reliable reward baselines, leading to stable gradients and better generalization but at higher memory and compute.
In SALT, this effect is amplified: the merge/diverge operations and step-level advantage rectification make the algorithm sensitive to group diversity. To investigate, we train GRPO+SALT on ALFWorld with Qwen2.5-1.5B-Instruct, testing group sizes $\{4,8,12,16\}$.
As shown in Figure~\ref{fig:group_size}, with a small group size of 4, GRPO slightly outperforms SALT, likely due to insufficient diversity for meaningful graph construction. However, as group size increases, SALT consistently surpasses the baseline, demonstrating its ability to leverage richer trajectory structure and confirming that larger groups enable more effective credit assignment through enhanced graph connectivity.

\section{Conclusion}
In this paper, we present SALT, a lightweight framework for assigning fine-grained, step-level advantages to improve group-based reinforcement learning for large language model agents. By constructing a trajectory graph that distinguishes shared and divergent steps across rollouts, SALT refines advantage assignment without requiring additional supervision, reward models, or architectural changes. As a plug-and-play module, it integrates seamlessly into existing RL pipelines such as GRPO and RLOO. Experiments on ALFWorld, WebShop, and AppWorld demonstrate consistent improvements across model sizes, highlighting SALT’s effectiveness and potential for future step-level RL methods.

\section*{Limitations}
Although SALT demonstrates strong empirical performance, several limitations remain.
First, while SALT is algorithm-agnostic in design, its application may require adaptation to specific environments. For benchmarks with discrete, well-structured action and state spaces, such as ALFWorld and WebShop, SALT can be applied directly with minimal preprocessing. However, in environments like AppWorld, where actions and observations reside in a continuous textual space, additional components, such as embedding models or clustering mechanisms, are necessary to meaningfully construct the trajectory graph.
Second, SALT introduces new hyperparameters, such as the history window length, which, while empirically not highly sensitive, still require tuning and may affect performance. 
Third, our current implementation integrates SALT only with RLOO and GRPO. Its compatibility with more advanced group-based policy optimization methods (e.g., DAPO~\citep{yu2025dapo}, GSPO~\citep{zheng2025group}) has not yet been explored.

\bibliography{custom}

\newpage
\appendix

\section{Related Work}
\textbf{LLM Agents.} The paradigm of LLMs has evolved beyond simple text generation to the development of LLM agents, which are autonomous systems designed to reason and perform complex, multi-step tasks. An agent's core functionality is centered on a central LLM that acts as its brain, enabling it to interpret a user's high-level goal, formulate a strategic plan, and execute that plan by interacting with its environment. This process typically involves a reasoning phase~\citep{yao2023react,shinn2023reflexion}, where the agent breaks down a task into actionable sub-goals; a planning phase~\citep{erdogan2025planandact,planning2}, where it determines the sequence of operations; and an action phase, where it leverages external tools, such as web search APIs~\citep{jin2025search,sun2025zerosearch}, code interpreters~\citep{codeact, feng2025retool}, or knowledge bases~\citep{zhu2025knowagent}, to gather information or manipulate data. By iteratively observing the results of its actions and adjusting its plan, an LLM agent demonstrates a significant step toward more generalized and autonomous problem-solving.

\textbf{Agent Training.} Early approaches for building agents primarily leveraged sophisticated prompting strategies and external tools to enhance performance on complex tasks, as exemplified by methods like ReAct~\citep{yao2023react}. However, models with smaller parameter counts often lack the requisite foundational capabilities for such complex reasoning. To address this limitation, some studies employ supervised fine-tuning (SFT) to enhance the models’ decision-making abilities~\citep{schick2023toolformer,zeng2024agenttuning,chen2023fireact}. More recently, there has been a growing focus on end-to end reinforcement learning for training agents~\citep{dong2025tool,singh2025agentic,chen2025r1-code}, which learn through direct, adaptive online interaction with an environment, thereby obviating the need for complex data preparation. Unlike supervised fine-tuning, RL naturally aligns with the objective of maximizing cumulative rewards through agent-environment interactions. This makes RL particularly well-suited for agentic tasks. Therefore, in this work, we focus on RL-based approaches to further enhance the capabilities of LLM agents.

\textbf{Step-level Supervision in RL.} In multi-step agent training, LLM agents must interact with the environment across multiple steps before receiving a final reward at the end of the trajectory~\citep{otc,chen2025r1-code}, making it challenging to assess the quality of intermediate actions. This lack of intermediate feedback hinders RL training and often requires extensive interactions with the environment to evaluate the trajectory. 
Incorporating step-level supervision or intermediate signals has been verified as effective in greatly enhancing the effectiveness and efficiency of LLM agent training~\citep{zhou2025sweet,wang2025spa,gigpo,choudhury2025process,wang2025harnessing}. 
However, existing methods either introduce additional information to guide the generation of step-level rewards~\citep{zhou2025sweet, gigpo, wang2025harnessing} or use another explicit process reward model to assign intermediate rewards~\citep{wang2025spa,choudhury2025process} which require deliberate designs and extra compute overhead.
As a result, in this paper, we propose a plug-and-play module which can be seamlessly integrated into existing RL algorithms to generate step-level feedbacks for multi-step training, without introducing any extra information or model.

\section{Experiment Details}
\label{app:exp}
% \subsection{Baselines}
% For WebShop and ALFWorld, the results of prompting methods are adopted from~\citet{gigpo}. The performance of PPO, RLOO, and GRPO training is evaluated based on our own implementation to ensure a fair comparison. 
% For AppWorld, the results of prompting methods and SFT are adopted from~\citet{loop}, and we report the results of RLOO training from our implementation. We refer readers to 
% \subsection{Training Details}
\subsection{Baselines}
For WebShop and ALFWorld, we adopt the results for prompting-based methods from~\citet{gigpo}. Performance of PPO, RLOO, and GRPO is evaluated using our own implementations to ensure fair and consistent comparisons under identical training conditions.

For AppWorld, we adopt the results for prompting-based methods and SFT-based methods from~\citet{loop}. Performance of RLOO and GRPO is evaluated using our own implementations to ensure fair and consistent comparisons under identical training conditions.
\subsection{Hyperparameters}
All models (versions with SALT and without SALT) are trained using the same hyperparameter configuration to isolate the effect of our proposed advantage refinement. Detailed settings are provided in Table~\ref{tab:hyper}.
\subsection{More Results}
A closer inspection of the per-difficulty results in Table~\ref{tab:app_diff} reveals that SALT consistently enhances performance on medium- and high-difficulty tasks (D2 and D3), while maintaining competitive results on easy tasks (D1). For example, on Test-N, GRPO+SALT improves TGC by +8.7pp on D2 (67.5\% → 76.2\%) and +6.1pp on D3 (35.5\% → 41.6\%), with even more substantial SGC gains—+11.3pp on D2 (38.7\% → 50.0\%) and +5.7pp on D3 (18.1\% → 23.8\%). On the more challenging Test-C, RLOO+SALT increases SGC on D3 from 7.7\% to 11.8\% (+4.1pp), and GRPO+SALT achieves a striking +10.6pp SGC improvement on D2 (8.0\% → 18.6\%). These trends confirm that SALT’s advantage-based credit assignment is particularly effective in complex, high-difficulty scenarios where precise action sequencing and long-horizon reasoning are required. Minor fluctuations on D1 (e.g., GRPO+SALT TGC on Test-C: 75.4\% → 70.3\%) are outweighed by consistent gains on harder tasks, underscoring SALT’s value in pushing the boundary of agent capabilities in realistic, challenging environments.

\begin{table}[t]
\centering
\resizebox{\columnwidth}{!}{
\begin{tabular}{l|ccc|ccc}
\toprule
\multirow{2}{*}{\textbf{Method}} & \multicolumn{3}{c|}{\textbf{Test-N TGC}} & \multicolumn{3}{c}{\textbf{Test-N SGC}}\\
& D1 & D2 & D3 & D1 & D2 & D3 \\
\midrule
RLOO & 81.6\textsubscript{\textpm1.9} & 63.0\textsubscript{\textpm6.8} & 37.7\textsubscript{\textpm3.4} & 59.2\textsubscript{\textpm6.8} & 31.2\textsubscript{\textpm7.6} & \textbf{22.6}\textsubscript{\textpm7.0} \\
\rowcolor{blue!10}RLOO+SALT & \textbf{81.7}\textsubscript{\textpm3.9} & \textbf{67.9}\textsubscript{\textpm4.6} & 37.7\textsubscript{\textpm4.2} & \textbf{67.3}\textsubscript{\textpm5.1} & \textbf{38.7}\textsubscript{\textpm6.1} & 14.3\textsubscript{\textpm4.2} \\
\midrule
GRPO & \textbf{85.3}\textsubscript{\textpm4.4} & 67.5\textsubscript{\textpm3.8} & 35.5\textsubscript{\textpm2.5} & 69.5\textsubscript{\textpm6.1} & 38.7\textsubscript{\textpm7.3} & 18.1\textsubscript{\textpm3.5} \\
\rowcolor{blue!10}GRPO+SALT & 84.9\textsubscript{\textpm3.0} & \textbf{76.2}\textsubscript{\textpm5.3} & \textbf{41.6}\textsubscript{\textpm2.3} & \textbf{72.6}\textsubscript{\textpm5.1} & \textbf{50.0}\textsubscript{\textpm8.8} & \textbf{23.8}\textsubscript{\textpm5.2} \\
\midrule
\multirow{2}{*}{\textbf{Method}} & \multicolumn{3}{c|}{\textbf{Test-C TGC}} & \multicolumn{3}{c}{\textbf{Test-C SGC}}\\
& D1 & D2 & D3 & D1 & D2 & D3 \\
\midrule
RLOO & 69.4\textsubscript{\textpm1.6} & 28.1\textsubscript{\textpm1.6} & 26.5\textsubscript{\textpm2.2} & 45.8\textsubscript{\textpm5.1} & 9.5\textsubscript{\textpm3.5} & 7.7\textsubscript{\textpm2.1} \\
\rowcolor{blue!10}RLOO+SALT & \textbf{74.4}\textsubscript{\textpm3.0} & \textbf{30.9}\textsubscript{\textpm3.3} & \textbf{28.5}\textsubscript{\textpm2.3} & \textbf{56.6}\textsubscript{\textpm4.9} & \textbf{10.4}\textsubscript{\textpm1.5} & \textbf{11.8}\textsubscript{\textpm3.0} \\
\midrule
GRPO & \textbf{75.4}\textsubscript{\textpm1.3} & 30.6\textsubscript{\textpm0.5} & 26.1\textsubscript{\textpm1.1} & \textbf{58.3}\textsubscript{\textpm3.3} & 8.0\textsubscript{\textpm0.0} & 8.7\textsubscript{\textpm0.7} \\
\rowcolor{blue!10}GRPO+SALT & 70.3\textsubscript{\textpm1.7} & \textbf{34.6}\textsubscript{\textpm0.5} & 26.1\textsubscript{\textpm3.2} & 48.6\textsubscript{\textpm3.9} & \textbf{18.6}\textsubscript{\textpm2.4} & \textbf{12.3}\textsubscript{\textpm4.3} \\
\bottomrule
\end{tabular}}
\caption{Detailed performance on AppWorld with respect to the task difficulty.}
\label{tab:app_diff}
\end{table}

\begin{table*}[h]
\centering
\vspace{-0in}
\resizebox{\textwidth}{!}{
\begin{tabular}{l|c|c|c|c}
\toprule
\multirow{2}{*}{Hyperparameter} & \multicolumn{2}{c|}{WebShop} & \multicolumn{2}{c}{ALFWorld} \\
 & \texttt{Qwen2.5-1.5B-Instruct}  & \texttt{Qwen2.5-7B-Instruct} & \texttt{Qwen2.5-1.5B-Instruct}  & \texttt{Qwen2.5-7B-Instruct} \\ \midrule
Mini-batch size      & 64 & 32 &  256  &128     \\ 
Max interaction steps      & \multicolumn{2}{c|}{15}     & \multicolumn{2}{c}{50}     \\ 
Max prompt length      & \multicolumn{2}{c|}{4096}     & \multicolumn{2}{c}{2048}     \\ 
State history length      & \multicolumn{4}{c}{3}     \\ 
Rollout temperature      & \multicolumn{4}{c}{1.0}     \\ 
Evaluation temperature      & \multicolumn{4}{c}{0.4}     \\ 
Group Size      & \multicolumn{4}{c}{8}     \\ 
Learning rate      & \multicolumn{4}{c}{1e-6}   \\ 
Max response length      & \multicolumn{4}{c}{512}   \\ 
KL loss coefficient      & \multicolumn{4}{c}{0.01}   \\ 
Training steps      & \multicolumn{4}{c}{300}   \\ 
Clip ratio   & \multicolumn{4}{c}{0.2}  \\
\bottomrule
\end{tabular}}
\caption{Hyperparameter settings used across all experiments for WebShop and ALFWorld.}
\label{tab:hyper}
\end{table*}

\begin{table*}[h]
\centering
\vspace{-0in}
\resizebox{0.5\linewidth}{!}{
\begin{tabular}{l|c}
\toprule
\multirow{2}{*}{Hyperparameter} & AppWorld  \\
 & \texttt{Qwen2.5-32B-Instruct} \\ \midrule
Mini-batch size      & 32  \\ 
Max interaction steps      &   40  \\ 
Max prompt length      &  28048  \\ 
State history length      &   3  \\ 
Rollout temperature      &  1.0 \\ 
Evaluation temperature      & 1.0 \\ 
Group Size      &   8 \\ 
Learning rate      & 1e-6 \\ 
Max response length      &  1500 \\ 
KL loss coefficient      & 0 \\ 
Training steps      & 100 \\ 
Clip ratio       & 0.2 \\
\bottomrule
\end{tabular}}
\caption{Hyperparameter settings used across all experiments for AppWorld.}
\label{tab:hyper-app}
\end{table*}

\section{Case Study}
To illustrate how SALT refines step-level advantages, we visualize four representative cases from ALFWorld. Each example shows the contextual state history, the current action step, and the transformation from original (trajectory-level) to updated (SALT-adjusted) advantage. These cases highlight SALT’s ability to distinguish between \textit{shared neutral steps} and \textit{distinct decisive steps} through trajectory graph analysis.

\begin{figure}[h]
\begin{tcolorbox}[width=\columnwidth,title = {Example 1: Shared Step $\to$ Advantage Averaged}]
\textbf{History state:}

-= Welcome to TextWorld, ALFRED! =-
You are in the middle of a room. Looking quickly around you, you see a cabinet 15, a cabinet 14, a cabinet 13, a cabinet 12, a cabinet 11, a cabinet 10, a cabinet 9, a cabinet 8, a cabinet 7, a cabinet 6, a cabinet 5, a cabinet 4, a cabinet 3, a cabinet 2, a cabinet 1, a coffeemachine 1, a countertop 4, a countertop 3, a countertop 2, a countertop 1, a drawer 4, a drawer 3, a drawer 2, a drawer 1, a fridge 1, a garbagecan 1, a microwave 1, a sinkbasin 1, a stoveburner 4, a stoveburner 3, a stoveburner 2, a stoveburner 1, and a toaster 1.

Your task is to: heat some cup and put it in cabinet.

\textbf{History step:}

go to countertop 1

\textbf{History state:}

You arrive at countertop 1. On the countertop 1, you see a butterknife 1, a papertowelroll 1, and a tomato 1.,

\rule{0pt}{1.5\baselineskip}

\textbf{Step:}

go to stoveburner 1
\tcblower
\textbf{Original advantage:} 1.007

\textbf{Updated advantage:} 0.008
\end{tcolorbox}
\caption{SALT reduces the advantage of this step because it frequently appears across both successful and failed rollouts. As a commonly executed transition (not task-determinative), its credit is averaged to avoid over-rewarding incidental actions.}
\end{figure}

\begin{figure}[h]
\begin{tcolorbox}[width=\columnwidth,title = {Example 2: Shared Step in Mixed Outcomes $\to$ Advantage Neutralized}]
\textbf{History state:}

You arrive at countertop 1. On the countertop 1, you see a fork 2, a ladle 1, a pan 1, a pot 1, a soapbottle 2, a spatula 1, a tomato 1, a window 3, and a window 1.

\textbf{History step:}

go to fridge 1

\textbf{History state:}

You arrive at fridge 1. The fridge 1 is closed.

\textbf{History step:}

open fridge 1

\textbf{History state:}

You open the fridge 1. The fridge 1 is open. In it, you see a potato 1.

\rule{0pt}{1.5\baselineskip}

\textbf{Step:}

go to countertop 2
\tcblower
\textbf{Original advantage:} -1.854

\textbf{Updated advantage:} 0.138
\end{tcolorbox}
\caption{Although this step originally received a large penalty (from a failed trajectory), SALT revises its advantage upward because it also appears in successful rollouts. This averaging prevents punishing actions that are contextually neutral or necessary.}
\end{figure}

\begin{figure}[h]
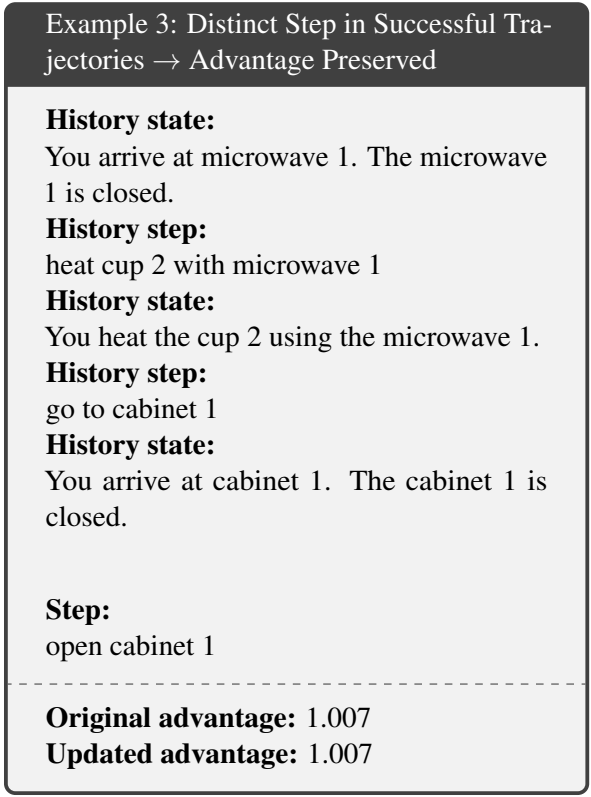

\begin{tcolorbox}[width=\columnwidth,title = {Example 3: Distinct Step in Successful Trajectories $\to$ Advantage Preserved}]
\textbf{History state:}

You arrive at microwave 1. The microwave 1 is closed.

\textbf{History step:}

heat cup 2 with microwave 1

\textbf{History state:}

You heat the cup 2 using the microwave 1.

\textbf{History step:}

go to cabinet 1

\textbf{History state:}

You arrive at cabinet 1. The cabinet 1 is closed.

\rule{0pt}{1.5\baselineskip}

\textbf{Step:}

open cabinet 1
\tcblower
\textbf{Original advantage:} 1.007

\textbf{Updated advantage:} 1.007
\end{tcolorbox}
\caption{SALT preserves the high advantage of this step because it occurs predominantly in successful trajectories and represents a task-critical action. No averaging is applied, as it is identified as a decisive, non-shared step.}
\end{figure}

\begin{figure}[h]
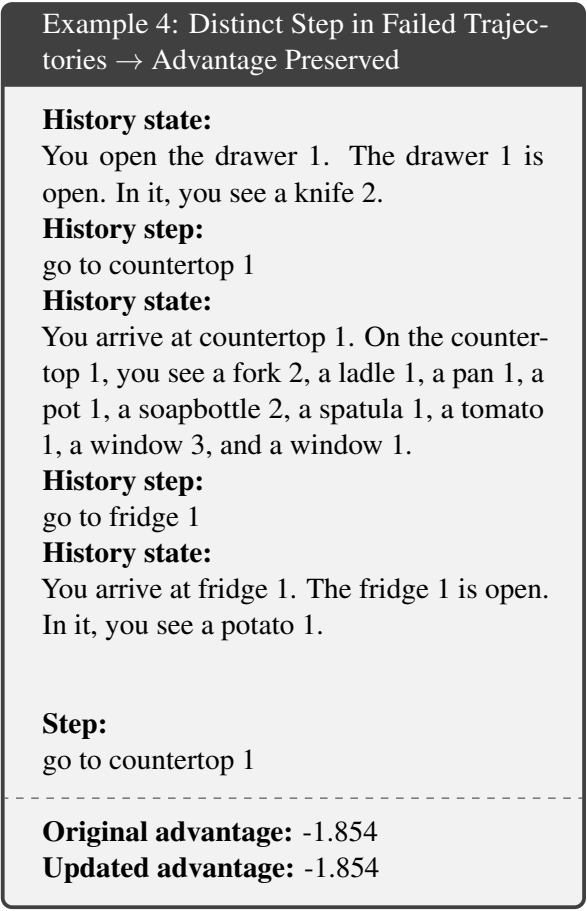

\begin{tcolorbox}[width=\columnwidth,title = {Example 4: Distinct Step in Failed Trajectories $\to$ Advantage Preserved}]
\textbf{History state:}

You open the drawer 1. The drawer 1 is open. In it, you see a knife 2.

\textbf{History step:}

go to countertop 1

\textbf{History state:}

You arrive at countertop 1. On the countertop 1, you see a fork 2, a ladle 1, a pan 1, a pot 1, a soapbottle 2, a spatula 1, a tomato 1, a window 3, and a window 1.

\textbf{History step:}

go to fridge 1

\textbf{History state:}

You arrive at fridge 1. The fridge 1 is open. In it, you see a potato 1.

\rule{0pt}{1.5\baselineskip}

\textbf{Step:}

go to countertop 1
\tcblower
\textbf{Original advantage:} -1.854

\textbf{Updated advantage:} -1.854
\end{tcolorbox}
\caption{SALT retains the negative advantage because this step appears primarily in failed trajectories and reflects inefficient backtracking. As a behaviorally distinct and detrimental pattern, it is not averaged with successful paths.}
\end{figure}

\end{document}